\documentclass[english, usemulticol]{cccconf}
\usepackage[comma,numbers,square,sort&compress]{natbib}
\usepackage{graphicx}
\usepackage{epstopdf}
\usepackage{amsmath}  
\usepackage{multirow}

\begin{document}

\title{Annotation and Detection of Emotion in Text-based Dialogue Systems with CNN}

\author{Jialiang Zhao\aref{bit},
        Qi Gao\aref{bit}}


\affiliation[bit]{School of Automation, 
        Beijing Institute of Technology, Beijing 100081, P.~R.~China
        \email{alanzjl@foxmail.com}
        \email{gaoqi@bit.edu.cn}}

\maketitle

\graphicspath{ {./pic/} }

\begin{abstract}
Knowledge of users' emotion states helps improve human-computer interaction. 
In this work, we presented EmoNet, an emotion detector of Chinese daily dialogues based on deep convolutional neural networks. 
In order to maintain the original linguistic features, such as the order, commonly used methods like segmentation and keywords extraction were not adopted, instead we increased the depth of CNN and tried to let CNN learn inner linguistic relationships. 
Our main contribution is that we presented a new model and a new pipeline which can be used in multi-language environment to solve sentimental problems.
Experimental results shows EmoNet has a great capacity in learning the emotion of dialogues and achieves a better result than other state of art detectors do.

\end{abstract}



\section{Introduction}

Emotion detection is potentially important in voice recognition systems, which provides improvements to further human-computer interactions.

Emotion detection can be achieved by various approaches, including text, gestures, facial expression, electrocardiographs, etc\cite{cohn1998,devillers2003,agrafioti2012,gunes2007}. 
Much work based on these features has been introduced over the last decade\cite{tinghao2009toward}.
However, text is the most common way for communication in the Internet. 
Text-based emotion detection has multiple applications in human-computer interaction, especially in search engines and chatting robots. 
With the knowledge of users' emotion, search engines could give more specific results, while chatting robots could reply in a more natural way.

Methods in text-based emotion detection can be divided into three categories: keyword-based detection, learning-based detection, and hybrid detection\cite{tinghao2009toward}. CNNs (convolutional neural networks) have shown their great capacity of learning the nature of images and speeches over the past few years. With a deep configuration and a large training dataset, keyword extraction might be not necessary anymore, which will significantly simplify the classification process.

In this work, we presented and tested a Chinese daily dialogues emotion detector based on deep convolutional neural network without segmentation or keyword extraction. 
This paper is divided into 7 sections. 
Section 2 gives an introduction of the pre-processing step.
Section 3 gives the CNN configuration of EmoNet.
Section 4 gives details of the training process.
In section 5, we tested different configurations and hyper-parameters and gave evaluations based on top-1 accuracy and comparisons with other state of art detectors in different scales.

\section{Dialogue Pre-processing}

We manually labeled over $12,000$ dialogues for training and evaluation.
They are divided into 4 categories: positive, negative, wondering, and neutral.
Most of them come from scripts of TV shows, movies and books.
Numbers of dialogues in different categories are listed in Tab.\ref{statistics}

\begin{table}[htb]
	\centering
	\caption{Numbers of dialogues in different categories}
	\label{statistics}
	\begin{tabular}{c|c|c|c|c}
		\hhline
		Overall     &  Positive	& Negative & Wondering & Neutral \\ \hline
		12186		&  3679		& 4205 & 1747 & 2555\\
		\hhline
	\end{tabular}
\end{table}

\subsection{Typical Routine}

Before diving into the training process, normally several pre-processing steps should be done in order to improve overall efficiency.
For English speech classification work, a typical routine is as follows:

\begin{itemize}
	\item Spell check: A widely used library for spell check is \emph{PyEnchant}. This step is optional.
	\item Stemming or Lemmatization\cite{stemming}: Both of them are used to find the original form of words, e.g. 'talk' is the original form of 'talking' or 'talked'.
	\item Case normalization: Convert input to uniform uppercase or lowercase. For example, "Today I'm happy." might be converted to "today i'm happy.".
\end{itemize}

However, for Chinese classification or recognition system, a special step, segmentation, generally needs to be done.
Normally researchers will build a LUT (look up table) which contains thousands of common used Chinese words before training, and then the frequency of appearance of each word in the target dialogue will be checked.
A number of segmentation algorithms have been developed over the last decade\cite{liu2002, peng2004, sproat1996}.
The frequency vector is finally used as training materials.

One of the most appealing advantage with this method is that outputs of segmentation are of uniform length (equals to the length of pre-built LUT), so that during most classification algorithms, e.g. KNN and SVM, it can achieve a better results.
Another important advantage is that segmentation with pre-selected keywords saves much time for reseachers and improves the efficiency\cite{huang2007}.

However, drawbacks with this algorithm shouldn't be ignored.

\begin{figure}[htb]
	\centering
	\includegraphics[width=200pt]{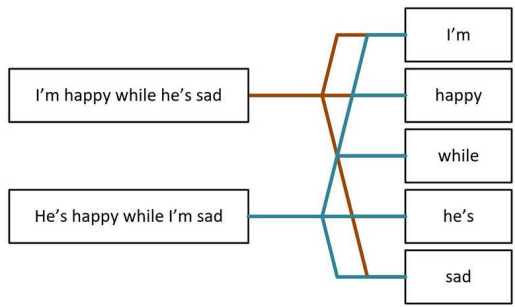}
	\caption{Loss of linguistic features in segmentation gives opposite emotion.}
	\label{overseg}
\end{figure}

\begin{itemize}
	\item Over segmentation: Similar to image precessing, over segmentation may significantly decrease the accuracy of our classifier\cite{overseg}. For example, 'not happy' may be segmented to 'not' and 'happy'. Obviously a single 'not' is meaningless but a 'happy' leads to a positive emotion, thus this sentence will be classified as positive because of over-segmentation, while the original emotion should be negative.
	\item Loss of linguistic features: Order of words is lost after segmentation. Same set of words with different orders sometimes gives absolutely opposite emotions. For example, as demonstrated in Fig.\ref{overseg}, "He's happy while I'm sad.", and "I'm happy while he's sad" share a same vocabulary, but give opposite emotion. 
\end{itemize}

\subsection{Pre-pocessing in EmoNet}

In order to avoid the problems listed above, we didn't adopt segmentation in EmoNet.
And because our net is tested in Chinese dialogues, stemming and lemmatization are not necessary.

\subsubsection{Removal of Stop Words}

Many words or expressions do not contribute to emotion or have same like-hood of occuring in those sentences not relevant to the target one. 
Such words are called stop words\cite{stopwords}. 
In order to save storage space and improve classification efficiency\cite{chinesestopwords}, we need to remove them before further training. 
A list of Chinese stop words from \emph{Data Hall} is used\cite{stopwordslist}. 

\begin{figure}[htb]
	\centering
	\includegraphics[width=200pt]{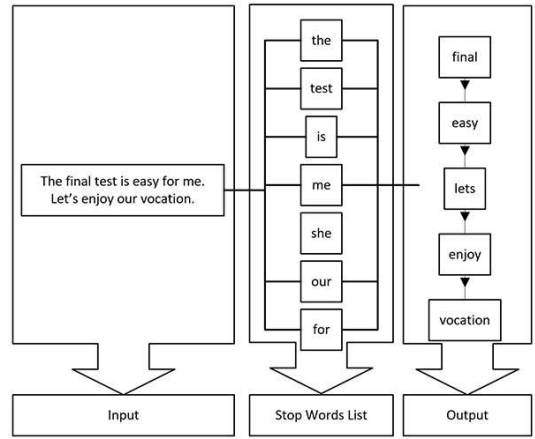}
	\caption{Removal of stop words. Meaningless and emotionless words are removed in this step. This figure gives a simple demonstration in English dialogues. This procedure works in a similar way with Chinese language. }
	\label{stopwords}
\end{figure}

As shown in Fig.\ref{stopwords}, we searched for stop words in every source dialogue and removed them in this step. Notice that the order is kept.

\subsubsection{Re-encode Input Characters}

UTF-8 (8-bit Unicode Transformation Format) is a large and powerful set of characters which encompasses most of the world's writing systems\cite{utf8}. 
With UTF-8, EmoNet gains the ability to perform in multi-language dialogue systems. 
Normally, one UTF-8 character is encoded with 1-6 byte, and in Chinese encoding, 3-byte UTF-8 is used.

However, because of UTF-8's large range, most information in this encoding system is useless. 
And because of the great number of Chinese characters, it's unnecessary to express each character precisely. 
The next step is to re-sample the encoding of input characters.

Firstly, because our net currently only contains Chinese, English and Arabic numerals, we can truncate UTF-8 to a new set that only contains characters of these vocabularies.
Tab.\ref{encoding} gives encoding of Chinese characters, full-width English characters and full-width Arabic numerals in UTF-8.

\begin{table}[htb]
	\centering
	\caption{UTF-8 encoding of full-width alphabets}
	\label{encoding}
	\begin{tabular}{l|l|l}
		\hhline
		Alphabet        &  Chinese	& Uppercase English \\ \hline
		UTF-8   	& u \emph{4e00} - u \emph{9fa5} 	& u \emph{ff21} - u \emph{ff3a}	\\ \hline
		Num			& 20902		& 26 \\ \hhline
		Alphabet    &  Lowercase English	& Numerals \\ \hline
		UTF-8   	& u \emph{ff41} - u \emph{ff5a} 	& u \emph{ff10} - u \emph{ff19}	\\ \hline
		Num			& 26		& 10 \\
		\hhline
	\end{tabular}
\end{table}

After this procedure, there are totally $20964$ characters left, most of which are Chinese characters.

\subsubsection{Encoding and Re-sampling}

Unlike image processing system, encoding of characters is not "continuous", which means adjacent codes not necessary give similar meanings, while value in imaging systems gives continuous effect in hue, saturation, etc.
Thus we don't need to represent each character precisely.
In EmoNet, we re-sample the $20964$ codes into $256$ codes with Eq.~\ref{eq:eq1}.

\begin{equation}
  \label{eq:eq1}
  \lambda_{new} = \lambda_{ori} \ \% \  256
\end{equation}

After this procedure, every character is represented with one Byte. 
We set the maximum length of one dialogue to be 144 in EmoNet, thus totally $144 \times 1 = 144 \ Bytes$ are used to represent one dialogue. 

\section{Convolutional Neural Network in EmoNet}

We adopted a deep CNN to do the classification job in EmoNet.
CNN is a powerful tool with an adjustable capacity by adjusting its depth and breadth.
With large datasets, CNN can achieve high performance in learning the nature of speech systems\cite{cnnintro}.

In EmoNet, we adopted a CNN which shares a similar but deeper configuration with LeNet-5\cite{lenet}, which consists of steps of convolutions, pooling, and fully-connected layers.

Fig.\ref{cnnstructure} gives the overall structure of EmoNet.

\singlecolumn{
\begin{figure}[htb]
	\centering
	\includegraphics[width=\hsize]{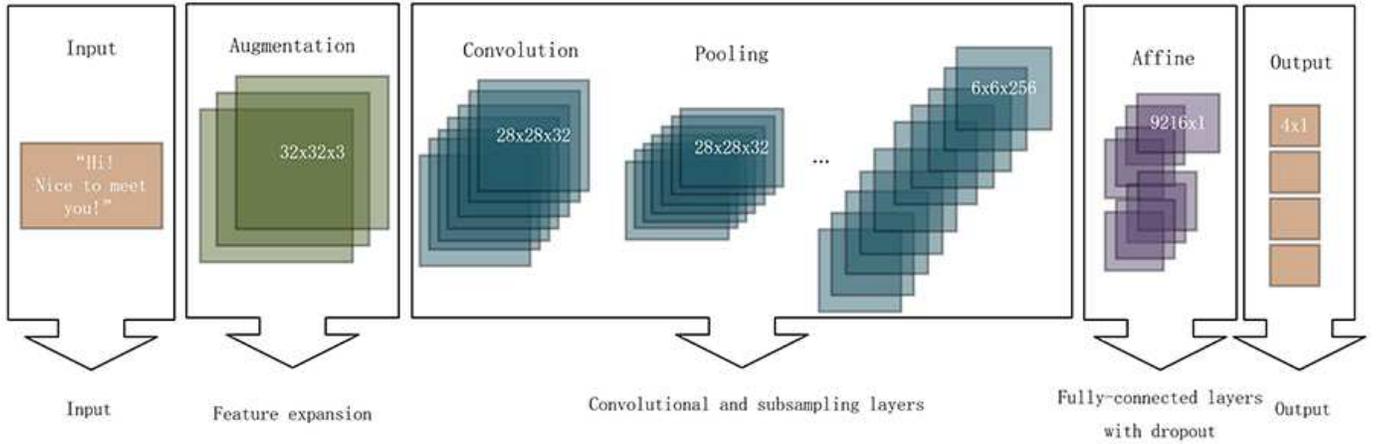}
	\caption{Network Structure of EmoNet. }
	\label{cnnstructure}
\end{figure}
}

Input layer is a $144\times1$ vector. The augmentation layer is used for feature expansion, which outputs a matrix of $32\times32\times3$. 
Five convolutional layers are used, and most of them come with a max pooling layer for sub-sampling and dimension reduction. 
The last convolutional layer gives an output of $6\times6\times256$. 
A fully-connected layer with dropout and \emph{softmax} method is used for classification.
The Output is of 4-dimension, positive, negative, wondering, and neutral.

Following sections give introduction of function and configuration of these layers in detail.

\subsection{Augmentation Layer}

We applied an augmentation layer to expand feature from the input and scale for the following layers.
Because generally every convolutional layer outputs a matrix of smaller size but larger dimension compared with its input matrix, in order to apply enough number of convolutional layers, we need to expand the input matrix to a larger square matrix. 

In every convolutional layer, we chose to use a filter of size $5\times5$ and stride $1$.
In order to outputs a $12\times12\times n$ in the last step of convolutional step and apply 5 convolutional layers, we need an square output of size 32 in the augmentation step, which can be calculated with Eq.~\ref{eq2}.

\begin{equation}
  \label{eq2}
  \begin{split}
  	Size &= S_{output} + N_{layers}\times(S_{filter} - Stride) \\
	&=12 + 5\times(5-1) \\
	&=32
  \end{split}
\end{equation}

An affine layer follows Eq.\ref{eq3}, where the weight matrix $W$ is of size $[144, 32\times32\times3=3072]$, and bias vector $B$ is of size $[3072, 1]$.

\begin{equation}
\label{eq3}
Output = W \cdot Input + B
\end{equation}

\subsection{Convolutional and Pooling Layer}

The convolutional part consists of 5 convolutional layers and 4 pooling layers.

\subsubsection{Convolutional Layer}

The convolutional layer adopts a filter to compute convolutions and generates a higher dimension matrix. In EmoNet, the filter sizes of all the 5 convolutional layers are chosen as $5\times5$. Filter dimensions vary in different layers.

Typically, a sigmoid nonlinearity (or activation function), which is given in Eq.\ref{sigmoid}, is used after every convolutional layer.

\begin{equation}
\label{sigmoid}
S(x) = \frac{1}{1+e^{-x}}
\end{equation}

However, because sigmoid is easy to saturate, and according to Nair and Hinton, we use RectifiedLinear Units (ReLUs)\cite{relu}, a non-saturating nonlinearity instead.
RELUs follow Eq.\ref{relu}.

\begin{equation}
\label{relu}
S(x) = \left\{
\begin{aligned}
x, \  x \geq 0 \\
0, \  x < 0
\end{aligned}
\right.
\end{equation}

\subsubsection{Pooling Layer}

Pooling layers introduce invariance, reduce dimension and prevent overfitting in CNNs\cite{pooling}.

We adopted overlapping pooling layers in this network. 
Overlapping pooling means adjacent pooling regions overlap with each other. 
For example, the pooling region size is denoted as $s\times s$, and stride is denoted as $z$. 
If $s \geq z$, which is used in most CNNs, it is non-overlapping. 
If $s < z$, then it becomes an overlapping pooling. 
According to Krizhevsky, Sutskever, and Hinton, an overlapping scheme will reduce error rate and decrease the probability of overfitting\cite{imagenet}. 
We adopted $s = 5, z = 1$ in EmoNet.

\begin{figure}[htb]
	\centering
	\includegraphics[width=150pt]{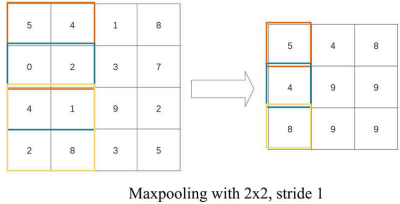}
	\caption{Maxpooling. Every $2\times2$ block from the original matrix is pooled into a single block with the max value. Dimension of output matrix can be gained with $D_{out} = \frac{D_{in} - (D_{filter} - Stride)}{Stride}$}.
	\label{maxpool}
\end{figure}

Also, many pooling methods have been developed, such as max pooling, average pooling, chunk pooling, etc. 
We used max pooling in EmoNet. 
Its principle is illustrated in Fig.\ref{maxpool}.

\subsection{Fully Connected Neural Network with Dropout}

\subsubsection{Fully Connected Layers}

A 3-layer fully connected neural network is connected to the last convolutional layer. Fully connected layers are affine layers, and every neuron in one layer is connected to each neuron in its adjacent layer. Each neuron of the first two layers comes with an RELU activation. Detailed configuration is listed in Tab.\ref{fullyconnect}.

\begin{table}[htb]
	\centering
	\caption{Structure of Fully Connected Layers}
	\label{fullyconnect}
	\begin{tabular}{l|l|l|l}
		\hhline
		Layer	& Layer 1     &  Layer 2	& Layer 3 \\ \hline
		Number of neurons   & 1024 	& 1024 & 5	\\ \hline
		Number of inputs	& 9216	& 1024 & 1024	\\ \hline
		Number of outputs	& 1024	& 1024 & 5	\\
		\hhline
	\end{tabular}
\end{table}

The final layer outputs a $5x1$ vector. This vector is then transformed into 5 categories, positive, negative, wondering, neutral and meaningless (which means this input should be discarded) by softmax classification.

\subsubsection{Overfitting}

Overfitting is a common problem in neural networks, especially when researchers don't have a large enough dataset. 
Nowadays there are two popular methods to prevent overfitting, batch normalization and dropout.

We tested EmoNet with batch normalization first. 
During training, inputs of every layer are always changing with the fact that parameters in the prior layers are changing at the same time. 
This phenomenon, which is also called internal covariate shift\cite{batchnorm}, slows down the training process and requires careful initialization.
Batch normalization solves this problem by normalizing layer inputs. 
However, in EmoNet, it reduced the error rate only by 2.5\% to 3.8\% compared with EmoNet without batch normalization with same amount of training. 
We think because our input is characters, which are discrete (while input of imaging systems are continuous), normalizing them might be unhelpful. 
Thus we tried dropout instead.

The core principle of dropout is to randomly disable some neurons, along with its connections during trainning\cite{dropout}, but do nothing during test. 
According to Srivastava et al., the optimal retention probability $p$ of input layer should be close to 100\% while with hidden layers it should be close to 50\%\cite{dropout}. 
We set $p_{input}$ as $1.0$ and $p_{hidden}$ as $0.3$, and we saw an reduction of error rate of around 6.5\%.

\section{Training Process}

Parameters need to be updated during training, and normally people use back-propagation\cite{backprop}. 
During this procedure, many parameters updating methods have been developed, such as SGD, momentum, RMSprop, and Adam. 
Adam is a first-order gradient-based optimization based on adaptive estimates of lower-order moments which can achieve a faster convergence than other methods like SGD, momentum and RMSprop\cite{adam}.

We chose Adam as EmoNet's parameter updating method. 
We divided the dataset into 2 parts, one of which was used for training, the other was used for evaluation. 
Cross validation\cite{crossvalidation} is used during the parameter updating process, and the training data was further divided into 32 mini-batch for cross validation.

\section{Optimization}

\subsection{Test of Different Configurations}
\begin{figure}[htb]
	\centering
	\includegraphics[width=\hsize]{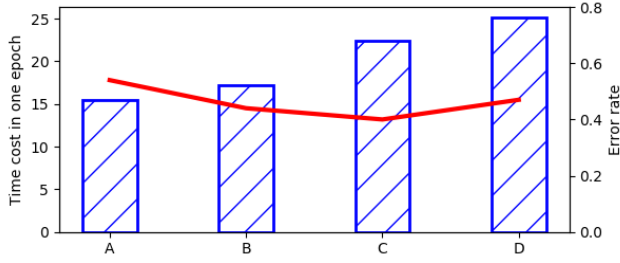}
	\caption{Test results of different configurations with same amount of data in one epoch.}
	\label{configtest}
\end{figure}

Configuration of EmoNet is modified from VGGNet\cite{vgg}. We tested different configurations of EmoNet. Validation error and training speed within one epoch using same amount of data are used as evaluation criteria. Tab.\ref{config} gives the configurations we tested, and Fig.\ref{configtest} gives comparison between them. Matplotlib\cite{matplotlib}, a python plotting package, is used for drawing.

We chose configuration B, which achieved relatively high accuracy and fast performance at the same time, as the final structure of EmoNet.

\subsection{Test of Different Parameters}

Learning rate, regulation strength in affine layers, and mean of initialization are also important hyper-parameters that give huge influence on final results. Because dropout makes careful initialization less important\cite{dropout}, we only tested different sets of learning rate $\gamma$ and regulation strength $L$. Results are collected with same amount of input data and same number of training iterations. Results are shown in Fig.\ref{paramtest}.

\begin{figure}[htb]
	\centering
	\includegraphics[width=180pt]{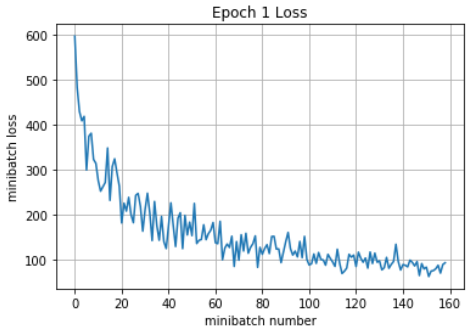}
	\caption{Loss v.s. time curve of the first epoch.}
	\label{curve}
\end{figure}

The group of highest accuracy, group $C$ is chosen, where $\gamma = 5\times10^{-6}, L = 1.5\times10^{-4}$. The loss v.s. time curve of the first epoch is shown in Fig.\ref{curve}.

\subsection{Evaluation}

With optimized parameters, EmoNet was trained with tensorflow and Nvidia GPU. After 100 epochs' training, an overall top-1 accuracy of 72.8\% was achieved. Tab.\ref{eval} gives accuracy of each category separately.

According to the results, neutral and wondering states are relatively easier to detect than neutral and negative states.

Wondering states are the easiest to detect, which is also easy to interpret. Generally dialogues of this emotion state come with question marks, or some specific words, like "what", "why", "how", etc.

\singlecolumn{
	\begin{table}[htb]
		\centering
		\caption{Configurations of EmoNet}
		\label{config}
		\begin{tabular}{c|c|c|c|c}
			\hhline
			No. &	A	& B     &  C	& D \\ \hline
			Number of layers &	9 weighted layers   & 9 weighted layers 	& 9 weighted layers & 9 weighted layers	\\ \hhline
			
			\multicolumn{5}{c}{Input $144\times1$ vector}	\\ \hline
			\multicolumn{5}{c}{Augmentation 3072 affine layer}	\\ \hline
			
			\multirow{2}{*}{ConvLayers 32-D} 
			& Conv32 & Conv32 & Conv32 & Conv32\\ 
			& Conv32 & 	& 	& \\ 
			\hline
			\multicolumn{5}{c}{Maxpool $5\times5$, stride $1$}	\\ \hline
			\multirow{2}{*}{ConvLayers 64-D} 
			& Conv64 & Conv64 & Conv64 & Conv64\\ 
			&  & Conv64 & 	& 	\\ 
			\hline
			\multicolumn{5}{c}{Maxpool $5\times5$, stride $1$}	\\ \hline
			\multirow{2}{*}{ConvLayers 128-D} 
			& Conv128 & Conv128 & Conv128 & Conv128\\ 
			& & & Conv128 & \\ 
			\hline
			\multicolumn{5}{c}{Maxpool $5\times5$, stride $1$}	\\ \hline
			\multirow{2}{*}{ConvLayers 256-D} 
			& Conv256 & Conv256 & Conv256 & Conv256\\ 
			& & & & Conv256 	\\ 
			\hline
			\multicolumn{5}{c}{Maxpool $5\times5$, stride $1$}	\\ \hline
			\multicolumn{5}{c}{Fully connected layer 1024}	\\ \hline
			\multicolumn{5}{c}{Dropout}	\\ \hline
			\multicolumn{5}{c}{Fully connected layer 1024}	\\ \hline
			\multicolumn{5}{c}{Dropout}	\\ \hline
			\multicolumn{5}{c}{Fully connected layer 5}	\\ \hline
			\multicolumn{5}{c}{Softmax}	\\ \hline
			\hhline
		\end{tabular}
	\end{table}
}

\singlecolumn{
\begin{figure}[htb]
	\centering
	\includegraphics[width=440pt]{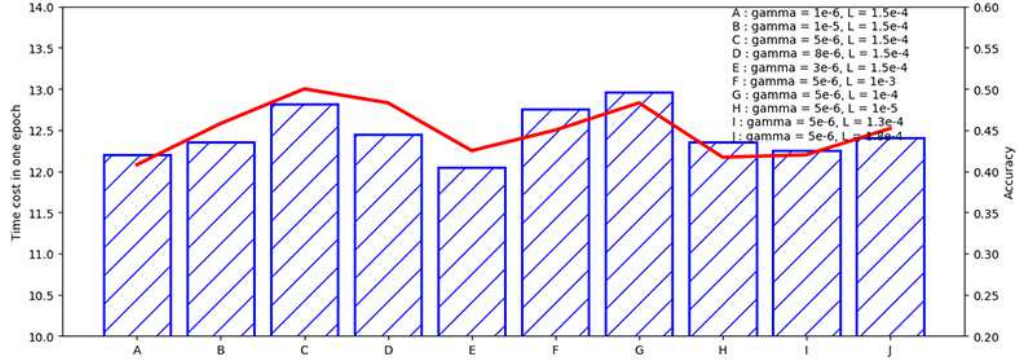}
	\caption{Test results of different parameters with same amount of data in one epoch.}
	\label{paramtest}
\end{figure}
}

Although accuracy of neutral state is high, we found that most emotion states of false detected dialogues were detected as neutral.
This phenomenon may result from two facts, the first one is that a neutral state is too vague and general, and the other one is that unlike other states in which we can find specific emotional words (for example, "why" for wondering states and "happy" for positive states), there are no specific words for neutral states.

Detection with positive states and negative states are similar. However, as we can see, accuracy with positive states is low. This may result from the quality of the training materials. Ironies are very common in our source, and it is sometimes even hard for us to distinguish whether a dialogue is positive or negative.

Comparisons between EmoNet and state of art Chinese text-based emotion detectors are given in Tab.\ref{compare}, Tab.\ref{compare_emotion}, and Tab.\ref{compare_neutral}.

\emph{Multi-Model Net} was given by Ze-Jing Chuang and Chung-Hsien Wu\cite{multimodal}, and \emph{ESiN} was given by Jianhua Tao\cite{esin}.

\begin{table}[!htb]
	\centering
	\caption{Evaluation}
	\label{eval}
	\begin{tabular}{l|l|l|l|l}
		\hhline
		Class		& neutral & positive& wondering& negative \\ \hline
		top-1 accuracy & 0.73& 0.57&  0.85&0.69		\\ 
		\hhline
	\end{tabular}
\end{table}

\begin{table}[!htb]
	\centering
	\caption{Comparison of overall accuracy}
	\label{compare}
	\begin{tabular}{l|l|l|l}
		\hhline
		Detector		& EmoNet & Multi-Modal Net& ESiN\\ \hline
		top-1 accuracy  & 0.72   & 0.6548	&  not given \\ 
		\hhline
	\end{tabular}
\end{table}

\begin{table}[!htb]
	\centering
	\caption{Comparison of accuracy of emotional states}
	\label{compare_emotion}
	\begin{tabular}{l|l|l|l}
		\hhline
		Detector		& EmoNet & Multi-Modal Net& ESiN\\ \hline
		top-1 accuracy  & 0.70   & 0.6466	&  over 0.7	\\ 
		\hhline
	\end{tabular}
\end{table}

\begin{table}[!htb]
	\centering
	\caption{Comparison of accuracy of neutral states}
	\label{compare_neutral}
	\begin{tabular}{l|l|l|l}
		\hhline
		Detector		& EmoNet & Multi-Modal Net& ESiN\\ \hline
		top-1 accuracy  & 0.73   & 0.7137	&  peak over 0.8 	\\ 
		\hhline
	\end{tabular}
\end{table}

\section{Conclusion}

In this paper, EmoNet was presented, an emotion detection system based on deep convolutional neural networks.
We analyzed the feature of Chinese encoding and adopted a pre-processing step without segmentation, stemming or lemmatization, which introduces difficulties but address the problem of loss of linguistic features.
Now in EmoNet, a simple re-sampling step was used to replace these steps.
In the future, we will try some other algorithms.
We are now working in developing a linear mapping system which can map arbitrary-length dialogues into equal-length outputs.

Different CNN configurations and different hyper-parameters were tested.
An overall accuracy of 0.72 was achieved with 12,000 training dialogues, 100 epochs' training and the optimized modal.
Top-1 accuracy of EmoNet is higher than other Chinese text based emotion detector.
However, it's foreseeable that with more training materials, EmoNet has the capacity to achieve a better performance.


\end{document}